\pdfoutput=1

\documentclass[11pt]{article}

\usepackage[final]{acl2023}
\usepackage{times}
\usepackage[T1]{fontenc}
\usepackage[utf8]{inputenc}
\usepackage{microtype}
\usepackage{inconsolata}
\usepackage{graphicx}
\usepackage{subcaption}
\usepackage{amsmath}
\usepackage{newtxmath}
\usepackage{listings}
\usepackage{booktabs,tabularx,array,xcolor,longtable}
\usepackage{placeins}
\newcommand{\factfb}[1]{\ensuremath{\mathrm{FactFB}_{#1}}}
\newcommand{\scorepairs}{\ensuremath{\mathrm{T\text{-}NLR}_{\mathrm{Score}}}}
\newcommand{\mlpairs}{\ensuremath{\mathrm{T\text{-}NLR}_{\mathrm{ML}}}}
\newcommand{\berttriple}{\ensuremath{\mathrm{BERT}_{\mathrm{triple}}}}
\hypersetup{
  pdftitle={Bridging Structure and Semantics: Knowledge Graph Verification via Triple-to-Text Alignment},
  pdfauthor={Ziang Cui},
  pdfkeywords={knowledge graphs, triple classification, natural language generation, triple-to-text alignment}
}
\title{Bridging Structure and Semantics: Knowledge Graph Verification via Triple-to-Text Alignment}
\author{Ziang Cui \\ Southeast University \\ \texttt{13776094815@163.com}}

\begin{document}
\maketitle
\thispagestyle{plain}
\raggedbottom

\begin{abstract}
Semantic accuracy is an important dimension of knowledge graph quality. Graph-based scoring methods exploit relational structure, while textual representations allow classifiers to draw on knowledge encoded by pre-trained language models (PLMs). We present GenVer, a generative verifier that scores triples using both their serialized form and generated natural language representations (NLRs). A triple--text alignment procedure selects training pairs for a sequence-to-sequence generator; a BERT-based classifier then combines features from each triple and its generated NLR. On four FactFB variants, constructed from FB15k using perturbed negative triples, GenVer obtains the highest classification accuracy among the evaluated baselines, ranging from 88.56\% to 91.52\%. The experiments suggest that generated representations can help triple classification, while the case analysis also exposes semantic drift and high-confidence errors. 
\end{abstract}

\section{Introduction}
Knowledge graphs (KGs) represent entities and their relations as structured assertions \cite{1}. Resources such as Freebase \cite{2}, DBpedia \cite{3}, and Wikidata \cite{4} support applications including social network analysis \cite{36}, recommendation \cite{35}, and question answering \cite{kbqa}. Errors in these resources can affect downstream applications, motivating the assessment of KG quality \cite{6}.

This work focuses on triple-level semantic accuracy: whether the assertion expressed by a triple is correct. We operationalize this objective as binary classification on a constructed dataset and assign each triple a score in $[0,1]$. A higher score indicates a stronger prediction of the positive class under the training distribution. Such a score is not, by itself, a calibrated probability that the assertion is true in the real world.

\begin{figure*}[t]
\centering
\begin{subfigure}[b]{0.48\textwidth}
\centering
\includegraphics[width=\linewidth]{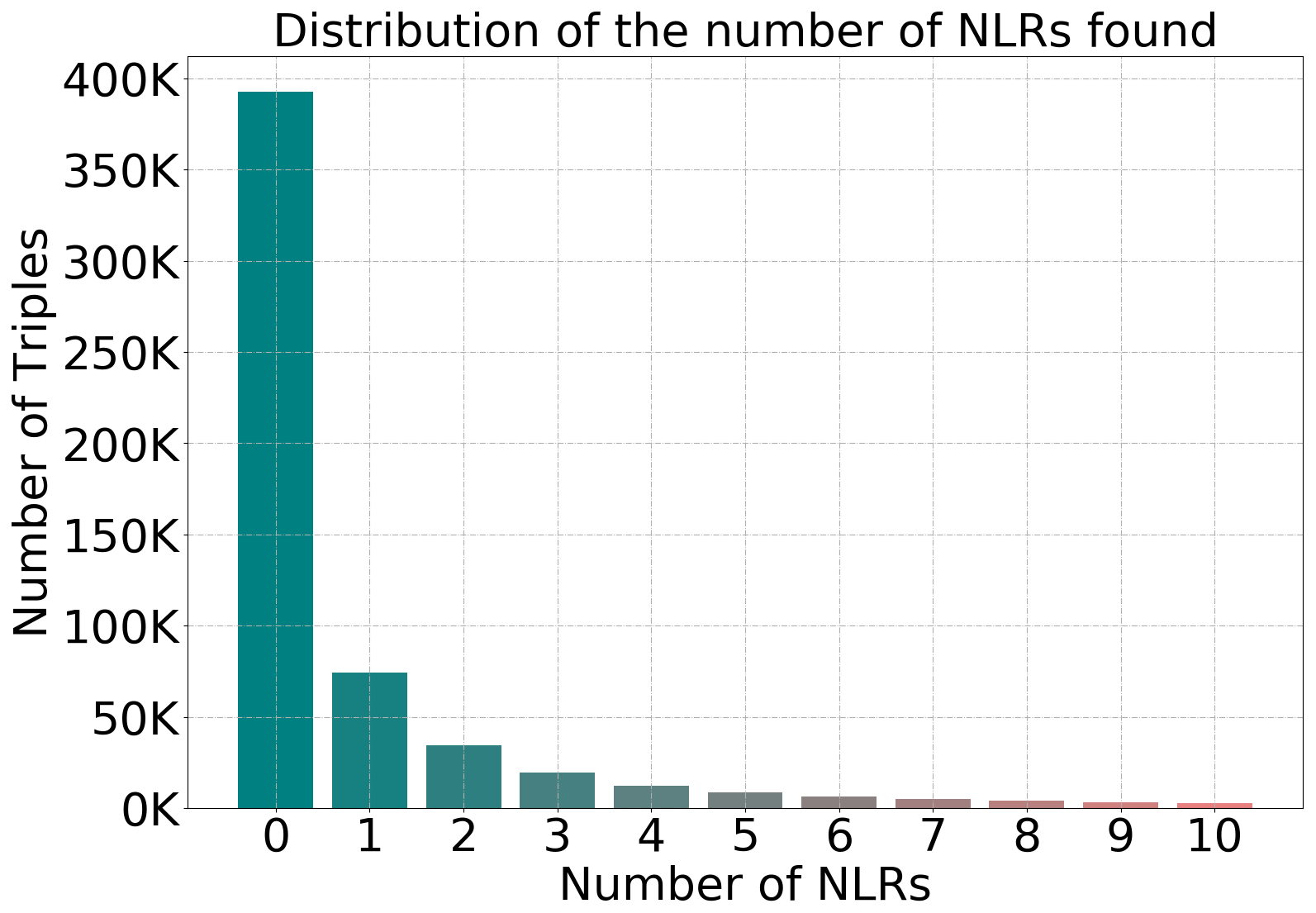}
\caption{Retrieval coverage.}
\end{subfigure}\hfill
\begin{subfigure}[b]{0.48\textwidth}
\centering
\includegraphics[width=\linewidth]{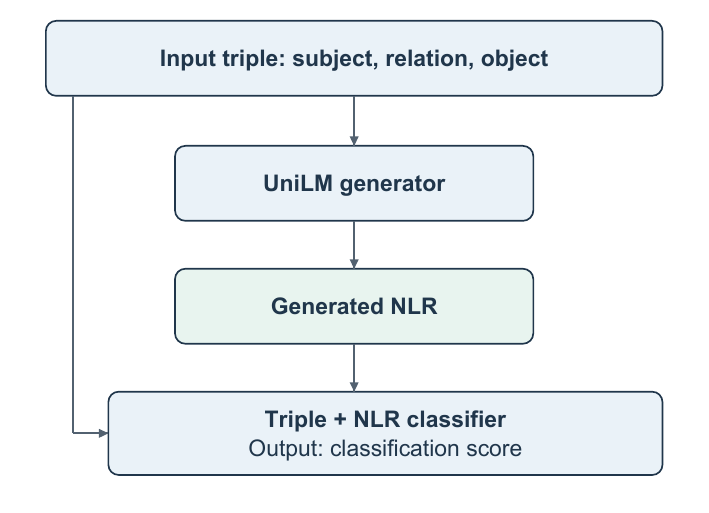}
\caption{Triple-to-text classification.}
\end{subfigure}
\caption{(a) Number of candidate NLRs retrieved per FB15k triple under the restricted search procedure. (b) GenVer uses generated text as a feature for triple classification.}
\label{fig:motivation}
\end{figure*}

Several established scoring methods learn entity and relation embeddings from graph structure \cite{16, 17, 19, 27}. Text-based verification methods additionally retrieve natural language evidence \cite{24, 25}. In this paper, a natural language representation (NLR) is a textual expression associated with a triple. We searched a Wikipedia dump\footnote{\url{https://dumps.wikimedia.org/enwiki}} for NLRs of FB15k triples. Figure~\ref{fig:motivation}(a) summarizes the results: many triples had no retrieved representation under our search procedure. This observation concerns the coverage of the restricted search described in Appendix~\ref{sparql}; it does not establish that relevant evidence is absent from Wikipedia or the Web.

GenVer generates an NLR for each input triple so that a classifier can use a textual representation even when the search procedure does not retrieve one. PLMs learn contextual representations from large text corpora \cite{8} and can encode factual associations \cite{9}. We use a sequence-to-sequence model to verbalize triples, then jointly encode the triple and its generated NLR for classification. Figure~\ref{fig:motivation}(b) illustrates this idea. Generated NLRs are model-produced features, not independent evidence for the truth of the input assertion; their faithfulness is an important limitation of the approach.

\textbf{Our contributions are:}
\begin{enumerate}
\item A two-stage framework that combines serialized triples and generated NLRs for triple classification.
\item A triple--text alignment procedure with four ranking features and two parameter-setting strategies for selecting generator training pairs.
\item A comparison on four FactFB variants, together with examples that illustrate useful representations and failures of generation or classification.
\end{enumerate}

\section{Related Work}
\subsection{Knowledge Graph Verification}
Methods for assessing triples broadly fall into two groups: representation learning and evidence retrieval.

\paragraph{Representation learning.}
These methods embed entities and relations in continuous vector spaces and learn scoring functions over triples. TransE \cite{16} models relations as translations, whereas DistMult \cite{28} and ComplEx \cite{19} use tensor factorization. KGTtm \cite{22} combines three estimators to assess triple trustworthiness. These approaches score triples directly or through a downstream classifier; graph-only configurations do not directly use textual descriptions.

\paragraph{Evidence retrieval.}
DeFacto \cite{24} uses search engines to retrieve web evidence for statements generated from triples using BOA patterns \cite{15}. FactCheck \cite{25} retrieves evidence from a static corpus and uses document features in a classifier to assess triples. Such methods introduce textual information into verification, but their coverage depends on the corpus and retrieval procedure. Missing or irrelevant evidence can limit their effectiveness.

\subsection{Knowledge Graphs and Pre-trained Language Models}
Work combining KGs and PLMs includes methods that use KGs to enhance language models and methods that use language models to process KGs.

\paragraph{KGs for PLMs.}
KgPLM \cite{14} incorporates factual knowledge through generative and discriminative pre-training tasks. Entities as Experts (EaE) \cite{18} learns representations for a fixed entity vocabulary and maintains an entity memory. FILM \cite{12} integrates symbolic knowledge into a neural language model through a fact memory. TEKGEN \cite{20} verbalizes Wikidata triples to construct a synthetic corpus for language model pre-training.

\paragraph{PLMs for KGs.}
KG-BERT \cite{kgbert} concatenates entity and relation names or descriptions and uses BERT to score triples for knowledge graph completion. SimKGC \cite{simkgc} uses textual information and contrastive learning with three types of negative examples to learn KG representations. GenVer also uses PLMs to process KGs, combining a serialized triple with a generated NLR for triple classification. Unlike retrieved evidence, the generated NLR is a model-produced feature and does not independently corroborate the triple.

\section{Proposed Method}
\begin{figure*}[htbp]
  \centering
  \includegraphics[width=1\linewidth]{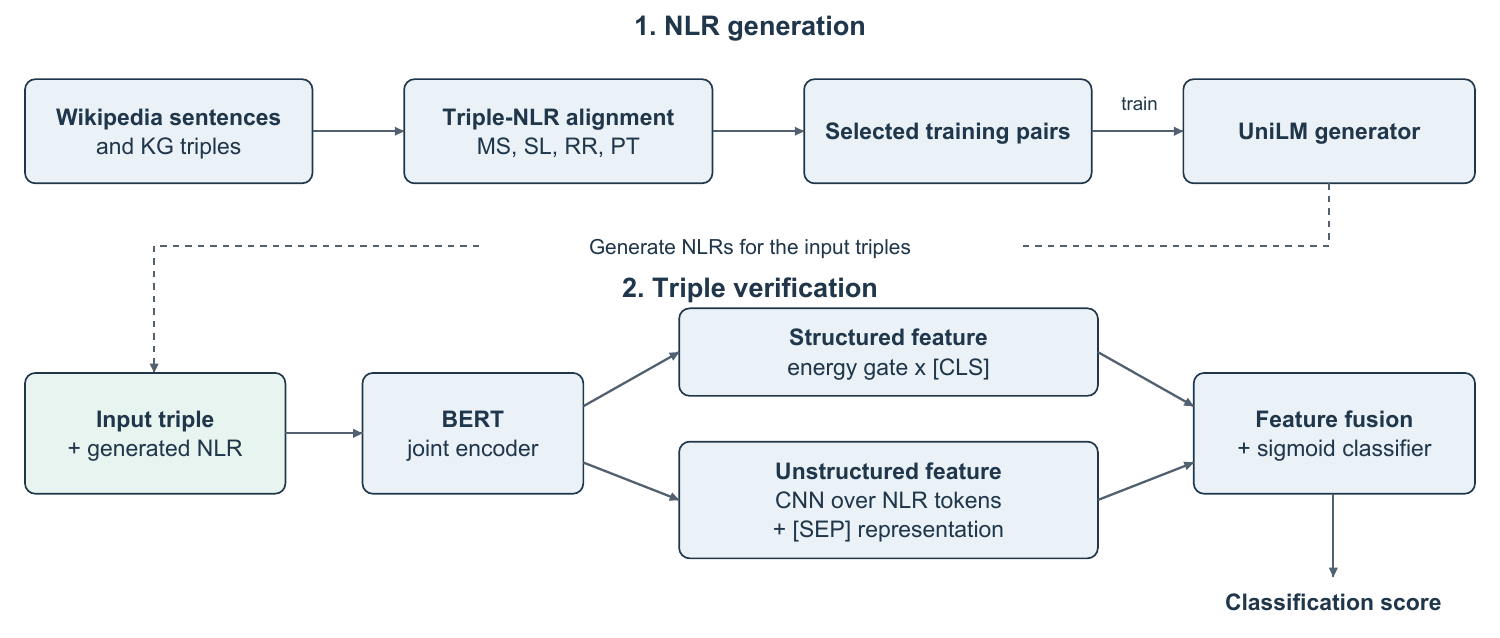}
  \caption{GenVer overview. Triple--NLR alignment supplies training pairs for the UniLM generator. At inference time, BERT jointly encodes each triple and its generated NLR; the two feature extractors feed a classifier. The stages are trained separately.}\label{data_flow}
\end{figure*}

GenVer consists of an NLR generation module and a triple verification module (Figure~\ref{data_flow}). First, a triple--NLR alignment procedure selects textual representations to train a generator. The trained generator then produces an NLR for each triple to be classified. Finally, a BERT-based verification module jointly encodes the serialized triple and its generated NLR, extracts features from both, and outputs a classification score.

\subsection{NLR Generation Module}
The NLR generation module selects triple--NLR training pairs and uses them to train a sequence-to-sequence model.

\paragraph{Triple--NLR alignment.}
Following the KG--text alignment approach of \citet{20}, we retrieve candidate sentences containing mentions of a triple's subject or object. Retrieval is restricted to the entity-associated Wikipedia pages described in Appendix~\ref{sparql}. A triple may match several sentences that do not faithfully express its relation. We therefore rank the candidates using four features: mention coverage (MS), sentence length (SL), relation relevance (RR), and pattern typicality (PT). All four features are computed with respect to the current triple.

\paragraph{Mention coverage (MS).} This feature considers whether an NLR contains mentions of both the subject and object of a triple. The feature for sentence $S$ is denoted as $\mathrm{MS}(S)$, where $\mathrm{MS}(S)=1$ if both mentions are present, and $\mathrm{MS}(S)=0$ otherwise.

\paragraph{Sentence length (SL).} Sentence length, $\mathrm{SL}(S)$, is the number of tokens in sentence $S$. The prior-based ranking rule penalizes longer sentences to discourage redundant information.

\paragraph{Relation relevance (RR).} $\mathrm{RR}(S)$ captures the relevance between a triple's relation and its NLR using string matching:
\begin{equation}
\mathrm{RR}(S) = \frac{\sum_{r \in L}\sum_{t \in S}\mathbf{1}[\operatorname{sim}(t,r)>\tau]}{|L|},\label{RR}
\end{equation}
where $L$ is the sequence of relation-label tokens, $\operatorname{sim}(t,r)$ is the Levenshtein ratio, $\tau$ is a similarity threshold, and $\mathbf{1}[\cdot]$ is the indicator function. The sum counts matching token pairs; the score is not constrained to $[0,1]$.

\paragraph{Pattern typicality (PT).} To identify common linguistic patterns, the typicality score for each token $t$ is calculated:
\begin{equation}
\operatorname{typ}(t) = \frac{\sum_{S\in\mathcal{N}}\mathbf{1}[t \in S]}{|\mathcal{N}|},
\end{equation}
where $\mathcal{N}$ represents the set of candidate NLRs for a triple. The pattern typicality feature $\mathrm{PT}(S)$ sums these scores:
\begin{equation}
\mathrm{PT}(S) = \sum_{t\in S}\operatorname{typ}(t).\label{PT}
\end{equation}

\paragraph{NLR ranking score.} We combine the four features into a ranking score $\mathrm{QS}$:
\begin{equation}
\mathrm{QS}(\mathbf{x}) = \operatorname{sigmoid}\!\left(\sum_{i=1}^{4}w_i x_i+b_s\right),\label{QS}
\end{equation}
where $w_i$ is the weight for each feature $x_i$, and $b_{s}$ is a bias term. $x_1$ through $x_4$ represent $\mathrm{MS}(S)$, $\mathrm{SL}(S)$, $\mathrm{RR}(S)$, and $\mathrm{PT}(S)$, respectively. We consider two parameter-setting strategies: a prior-based rule and supervised learning from labeled pairs.

For each triple with retrieved candidates, we select the NLR with the highest $\mathrm{QS}$ to form the generator training corpus. Section~\ref{alignment_experiment} compares the downstream classification results obtained with the two parameter-setting strategies.

\paragraph{NLR generation.}
We use UniLM \cite{26} as the sequence-to-sequence generator. Each input triple is serialized as \texttt{[CLS]} followed by its subject, relation-label tokens, and object. Entities are added to the model vocabulary as individual tokens rather than represented by multi-token labels.

\subsection{Triple Verification Module}
BERT \cite{8} jointly encodes each triple and its generated NLR. The input consists of \texttt{[CLS]}, the subject, relation-label tokens, the object, \texttt{[SEP]}, and the NLR tokens, in that order. Two extractors then derive features from the contextualized token representations.

\paragraph{Unstructured feature extractor.}
Let $\mathbf{X}\in\mathbb{R}^{m\times d}$ contain the BERT representations of the $m$ NLR tokens, with hidden dimension $d$. Following \citet{29}, a convolutional filter $\mathbf{W}\in\mathbb{R}^{h\times d}$ with window size $h$ computes
\begin{equation}
c_i = f\!\left(\langle\mathbf{W},\mathbf{X}_{i:i+h-1}\rangle+b\right),\label{conv}
\end{equation}
where $\langle\cdot,\cdot\rangle$ denotes the elementwise inner product, $b$ is a bias, and $f$ is an activation function. Global max pooling yields
\begin{equation}
\hat{c} = \max_i c_i.\label{max}
\end{equation}
Filter banks with three window sizes produce vectors $\mathbf{z}_1$, $\mathbf{z}_2$, and $\mathbf{z}_3$, which are concatenated as $\mathbf{h}=[\mathbf{z}_1;\mathbf{z}_2;\mathbf{z}_3]$. Let $\mathbf{p}_{\mathrm{sep}}$ be the BERT representation of \texttt{[SEP]}. The unstructured feature is
\begin{equation}
\mathbf{u} = \mathbf{W}_{u}[\mathbf{h};\mathbf{p}_{\mathrm{sep}}] + \mathbf{b}_u.
\end{equation}

\paragraph{Structured feature extractor.} Inspired by TransE \cite{16}, we use an energy-based gate. Let $\mathbf{e}_s$, $\mathbf{e}_p$, and $\mathbf{e}_o$ be the subject, relation, and object representations from BERT. The energy $E$ is:
\begin{equation}
E = \Vert \mathbf{e}_s+\mathbf{e}_p-\mathbf{e}_o \Vert_2.
\end{equation}
The scalar gate $a$ modulates the triple feature:
\begin{equation}
a = \frac{1}{1+e^{\lambda(\theta-E)}},
\end{equation}
where $\lambda$ and $\theta$ are trainable parameters. The trainable parameter $\lambda$ determines whether the gate increases or decreases with energy. The gate is a feature weight, not a truth probability. The final structured feature is $\mathbf{s}=a\,\mathbf{l}$, where $\mathbf{l}$ is the \texttt{[CLS]} token representation. Here, ``structured'' refers to the subject--relation--object organization of the BERT input rather than a separate graph-neighborhood encoder.

\paragraph{Classification score.}
The two feature vectors are concatenated and projected as $\mathbf{t}=\mathbf{W}_t[\mathbf{u};\mathbf{s}]+\mathbf{b}_t$. Here and above, semicolons denote concatenation. A final sigmoid produces the score
\begin{equation}
o = \operatorname{sigmoid}(\mathbf{w}_o^{\top}\mathbf{t}+b_o),\label{output}
\end{equation}
where $o\in[0,1]$ is a classifier output. Classification accuracy and the area under the receiver operating characteristic curve (AUROC) measure agreement with dataset labels, not probability calibration.

\section{Experiments}
\subsection{Datasets}
\textbf{FactFB:} We use FB15k \cite{16} as the source of positive triples and select its 2, 4, 8, and 16 most frequent relations to construct four FactFB variants. Negative triples are generated by randomly perturbing an entity or a relation in a positive triple. Perturbed triples that already occur in the knowledge graph are excluded from the negative examples. These variants, denoted \factfb{2}, \factfb{4}, \factfb{8}, and \factfb{16}, differ in size and relation coverage but are not independent domains. Appendix~\ref{dataset_info} gives dataset statistics. A perturbed triple is a constructed negative example; absence from a KG alone does not establish that an assertion is false in the real world.

\paragraph{Triple--NLR pairs.}
We apply the alignment procedure to a Wikipedia dump, using entity-page mappings to restrict retrieval (Appendix~\ref{sparql}). Two generator training corpora are constructed by setting the ranking parameters either from prior assumptions or from manually labeled pairs.

For the prior-based strategy, the default parameters in Eq.~\eqref{QS} are $w_1=2.224, w_2=-0.088, w_3=5.392, w_4=3.803,$ and $b_s=-1.342$ as derived in Appendix~\ref{app:default-parameters}. We denote the resulting corpus by \scorepairs{}.

For the learned strategy, we manually label the candidate pairs: label 1 indicates that the NLR fully expresses the triple, and label 0 indicates that it does not. The annotation set contains 60 pairs for fitting and evaluating the ranker in Eq.~\eqref{QS}. The ranker was fitted on 48 pairs and evaluated on the remaining 12 pairs (a 20\% held-out split), with a reported accuracy of 100\%. This small sample limits conclusions about generalization. We fit logistic regression using cross-entropy loss and use its positive-class probability as the ranking score. The learned parameters are $w_1=2.603, w_2=-0.015, w_3=-0.427, w_4=-0.402,$ and $b_s=-1.218$. This dataset, produced using the learned parameters, is denoted by \mlpairs{}.

\subsection{Effect of Training-Pair Selection}\label{alignment_experiment}
We compare GenVer trained with the two triple--NLR corpora. Following KGTtm \cite{22}, we evaluate binary triple classification. The model settings are provided in Appendix~\ref{settings}.

\begin{table}[tbp]
  \caption{Classification accuracy (\%) using different triple--NLR pairs.}
  \label{ML_Score}
  \centering
\small
\setlength{\tabcolsep}{2.5pt}

  \begin{tabular*}{\linewidth}{@{\extracolsep{\fill}}lrrrr@{}}
\toprule
    Corpus    & \factfb{2}  & \factfb{4}& \factfb{8}& \factfb{16}\\
\midrule
    \scorepairs{} &      86.12  &   88.28&   86.55  & 88.35 \\
   \mlpairs{}   &     \textbf{91.52}   &  \textbf{90.67} &    \textbf{88.56}  & \textbf{90.51}\\ 
\bottomrule
\end{tabular*}
\end{table}

\begin{figure*}[t]\centering
\begin{subfigure}{0.47\textwidth}\centering
\includegraphics[width=\linewidth]{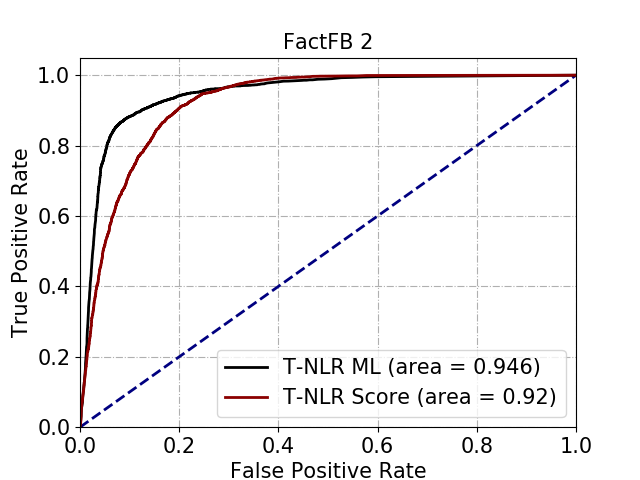}
\caption{\factfb{2}.}\end{subfigure}\hfill\begin{subfigure}{0.47\textwidth}\centering
\includegraphics[width=\linewidth]{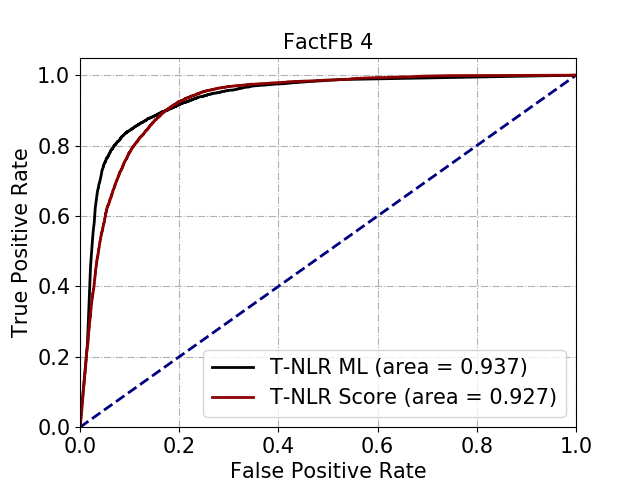}
\caption{\factfb{4}.}\end{subfigure}\par\medskip
\begin{subfigure}{0.47\textwidth}\centering
\includegraphics[width=\linewidth]{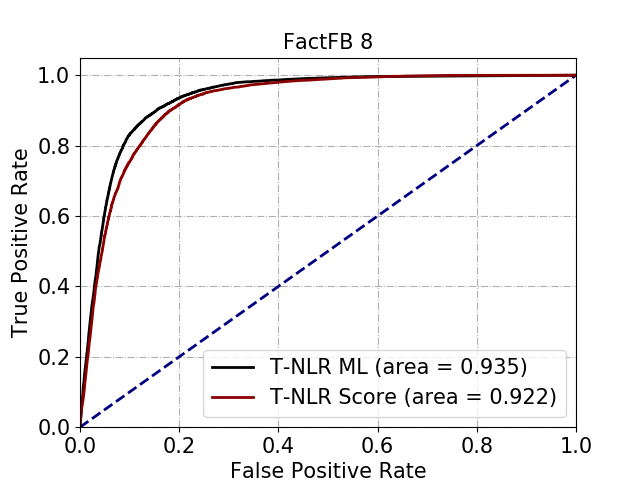}
\caption{\factfb{8}.}\end{subfigure}\hfill\begin{subfigure}{0.47\textwidth}\centering
\includegraphics[width=\linewidth]{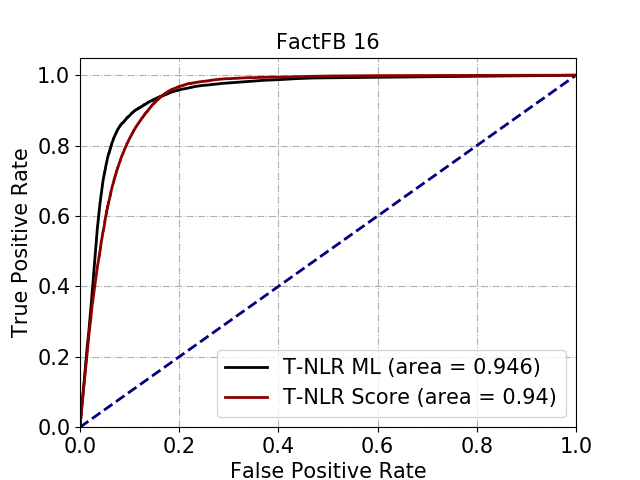}
\caption{\factfb{16}.}\end{subfigure}
\caption{ROC curves for GenVer trained with \scorepairs{} and \mlpairs{}. AUROC values are shown in the legends.}\label{roc}
\end{figure*}

Table~\ref{ML_Score} reports higher classification accuracy for \mlpairs{} than for \scorepairs{} on all four FactFB variants. This comparison changes the generator training pairs and measures the resulting downstream performance; it is not a direct evaluation of NLR faithfulness.

Figure~\ref{roc} reports ROC curves and AUROC values for the two training-pair variants. An ROC curve plots the true-positive rate against the false-positive rate as the decision threshold changes; AUROC summarizes ranking performance across thresholds. The reported AUROC values are higher with \mlpairs{} on each FactFB variant. These observations associate the training-pair selection strategy with downstream performance, without isolating which properties of the generated text account for the difference.

Appendix~\ref{app:generated-nlrs} presents generated examples from the evaluated relations. These examples include omissions, unsupported additions, and changes to relation meaning, so shorter or more fluent outputs should not automatically be interpreted as more faithful representations.

The learned ranking configuration performs better in these experiments. The small annotation sample used to fit it, however, limits conclusions about the reliability of this selection procedure beyond the evaluated setting.

\subsection{Comparison with Existing Methods}
\begin{table}[tbp]
  \centering
  \caption{Triple classification accuracy (\%).}
  \label{acc}
  \centering
\small
\setlength{\tabcolsep}{2.5pt}

  \begin{tabular*}{\linewidth}{@{\extracolsep{\fill}}lrrrr@{}}
\toprule
    Methods    & \factfb{2}  & \factfb{4}& \factfb{8}& \factfb{16}\\
    \midrule
GenVer   &     \textbf{91.52}   &  \textbf{90.67} &    \textbf{88.56}  &  \textbf{90.51}   \\ 
DistMult &      81.63  &   79.24&   81.17  & 83.33\\
ComplEx &     84.46  &   89.68&   87.20  & 89.83\\
TransE &      87.39  &   86.20&     85.27& 85.63\\
TransH &      88.73  &   85.24&     83.80& 83.31\\
TransR &      84.43    &   69.98&   62.28 & 59.80\\
TransD &      68.95  &   67.35&     64.55& 54.11\\
KGTtm &      80.94  &    86.76  & 85.97  & 88.46    \\
\berttriple{} &      85.33  &    88.84  & 88.53  &   88.06  \\
SimKGC &      82.78  &    78.62  & 80.26  &   85.47  \\
    \bottomrule
\end{tabular*}
\end{table}

We compare GenVer with the graph embedding methods TransE \cite{16}, ComplEx \cite{19}, DistMult \cite{28}, TransH \cite{32}, TransR \cite{33}, and TransD \cite{34}, using OpenKE \cite{30}. We also include KGTtm \cite{22}, which combines several estimators for triple evaluation. \berttriple{} is an ablation that omits the generated NLR and its corresponding feature extractor. SimKGC \cite{simkgc} provides a text-based KG representation baseline. The training settings are listed in Appendix~\ref{settings}. These comparisons concern the implemented configurations on FactFB, rather than an exhaustive comparison with current verification systems.

Table~\ref{acc} reports triple classification accuracy. GenVer has the highest accuracy in each column. Its margins over the strongest other baseline are 2.79, 0.99, 0.03, and 0.68 percentage points on \factfb{2}, \factfb{4}, \factfb{8}, and \factfb{16}, respectively. In particular, the difference on \factfb{8} is small (88.56 versus 88.53). Run-to-run variability and statistical significance are not reported, so the margins alone do not establish a reliable advantage.

\begin{figure}[tbp]
  \centering
  \includegraphics[width=\linewidth]{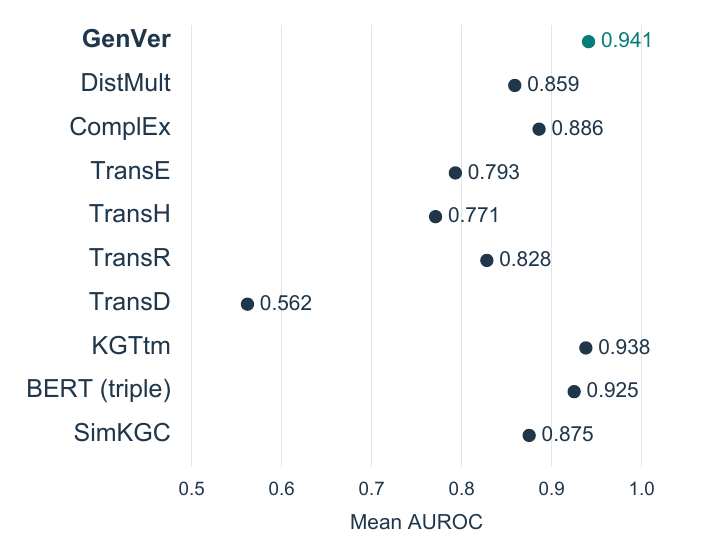}
  \caption{Mean AUROC across the four FactFB variants, for the evaluated model configurations.}\label{AUC}
\end{figure}

The relative performance of the baselines varies across the four relation subsets. Figure~\ref{AUC} summarizes their mean AUROC, with GenVer reaching 0.941. Together with the accuracy results, this supports the usefulness of generated text features in the evaluated configuration, but does not establish robustness across domains, calibrated factual confidence, or superiority under a different negative-sampling protocol.

\subsection{Case Study}
We examine examples from two frequent relations, \textit{Film Date Region} and \textit{Award Nomination}, to illustrate how generation quality and classification confidence can diverge. For each relation, we report one true positive (TP), true negative (TN), false positive (FP), and false negative (FN), with correctness defined by the dataset labels. Relation names are abbreviated for readability.

\begin{table*}[t]
\centering
\caption{Examples for \textit{Film Date Region}. Target denotes the dataset label; score denotes the classifier output. Generated wording is preserved verbatim.}
\label{film}
\small
\setlength{\tabcolsep}{4pt}
\renewcommand{\arraystretch}{1.13}
\begin{tabularx}{\textwidth}{@{}>{\raggedright\arraybackslash}p{2.35cm}>{\raggedright\arraybackslash}p{2.5cm}>{\raggedright\arraybackslash}Xcrl@{}}
\toprule
Subject & Object & Generated NLR & Target & Score & Case \\
\midrule
Remember the Titans & United States of America & Remember the Titans opened strongly at the United States of America box office & 1 & 0.9999993 & TP \\
\addlinespace[4pt]
Bruce Cohen & United States of America & Bruce Cohen was ranked second in the United States of America & 0 & 0.0000003 & TN \\
\addlinespace[4pt]
The Grudge & Buenos Aires & ``I finally found someone" by Buenos Aires and Bryan Adams & 0 & 0.9997579 & FP \\
\addlinespace[4pt]
Wild at Heart & France & Wild at Heart released the film in France & 1 & 0.0203743 & FN \\
\bottomrule
\end{tabularx}
\vspace{12pt}
\caption{Examples for \textit{Award Nomination}, using the same conventions as Table~\ref{film}.}
\label{award}
\small
\setlength{\tabcolsep}{4pt}
\renewcommand{\arraystretch}{1.13}
\begin{tabularx}{\textwidth}{@{}>{\raggedright\arraybackslash}p{2.35cm}>{\raggedright\arraybackslash}p{2.5cm}>{\raggedright\arraybackslash}Xcrl@{}}
\toprule
Subject & Object & Generated NLR & Target & Score & Case \\
\midrule
MC Hammer & Grammy Award & Hammer earned a Grammy Award for Album of the Year at the 31st Grammy Awards. & 1 & 0.9999999 & TP \\
\addlinespace[4pt]
Raymond Burr & Blake Edwards & Antarctica1 has been nominated for the Blake Edwards & 0 & 0 & TN \\
\addlinespace[4pt]
Geoffrey Rush & Golden Raspberry Award & Which featured Grammy Award for Best Boxed or Special Limited Edition Package sold over 7.6 million records worldwide & 0 & 0.9889689 & FP \\
\addlinespace[4pt]
Anne Hathaway & MTV Movie Award & Entertainment Weekly ranked MTV Movie Award for Best Fight the fifth-greatest film of all time & 1 & 0.0642089 & FN \\
\bottomrule
\end{tabularx}
\end{table*}

Table~\ref{film} reproduces four examples for \textit{Film Date Region}, with dataset labels and classifier scores. The high-scoring positive example associates a film with a box-office region, but does not explicitly specify a release event or date. The low-scoring negative example uses a person as the subject of a film-release relation, illustrating a type mismatch that may make the constructed task easier. These examples do not separately identify the contribution of graph structure, language-model knowledge, and NLR features.

The false positive is more revealing: the input about \textit{The Grudge} is mapped to an unrelated sentence about a song, yet receives a score of 0.9997579. This is a failure to reject the dataset's negative example despite an unfaithful NLR. It does not demonstrate that the generated sentence itself is factual or establish why the classifier assigns a high score. Conversely, the \textit{Wild at Heart} example receives a low score despite its positive dataset label. Without independent evidence, that prediction is not a reason to revise the label or prefer another release region.

Table~\ref{award} shows related failures for \textit{Award Nomination}. The high-scoring example about MC Hammer is classified correctly according to the dataset label, but its NLR strengthens a nomination relation into a claim about winning an award and adds a category and ceremony. Correct classification therefore does not imply faithful generation. The Geoffrey Rush example receives a high score despite omitting the subject and changing the award context; the Anne Hathaway example receives a low score with another unfaithful NLR. These cases motivate separate assessment of representation faithfulness and classification correctness.

The cases illustrate that generated NLRs can be useful features but can also omit entities, change relations, or introduce unsupported details. High confidence can accompany an incorrect prediction. They are qualitative examples, not a quantitative demonstration of a causal relationship between fluency and classifier confidence.

\FloatBarrier
\section{Conclusion}
We presented GenVer, a framework that combines serialized triples with generated natural language representations for triple classification. A triple--text alignment procedure supplies training pairs for a generator, and a BERT-based classifier combines triple and NLR features. The FactFB experiments favor GenVer over the evaluated baselines in classification accuracy, with margins that vary substantially across subsets. Generated representations also exhibit semantic drift, and the classifier can make high-confidence errors. The findings are therefore limited to the constructed classification setting; faithful generation and evidence-based verification remain unresolved.

\section{Limitations}
The evaluation uses four subsets of FB15k with randomly perturbed negative examples. Such negatives can contain obvious type mismatches and may not represent naturally occurring KG errors. The four variants do not establish cross-domain generalization, and a missing or perturbed edge is not necessarily false under an open-world interpretation.

Generated NLRs are not independent evidence for a triple. They can omit entities, alter relations, or add unsupported facts, as shown in the case study and appendix. The generation and classification stages are trained separately, and downstream accuracy does not directly measure NLR faithfulness.

\begingroup
\small
\bibliography{refs}
\bibliographystyle{acl_natbib}
\endgroup

\appendix
\clearpage
\onecolumn
\section{Supplementary Details}
\subsection{Dataset Statistics}
\label{dataset_info}
Table~\ref{FactFB} summarizes the entity, relation, and triple counts for the four FactFB variants.
\begin{table}[ht]
\centering
\caption{Sizes of the FactFB variants.}
\label{FactFB}
\begin{tabular}{lrrrr}
\toprule
Property & \factfb{2} & \factfb{4} & \factfb{8} & \factfb{16} \\
\midrule
Entities & 4085 & 5680 & 7230 & 7621 \\
Relations & 2 & 4 & 8 & 16 \\
Triples & 142088 & 260640 & 460464 & 752164 \\
\bottomrule
\end{tabular}
\end{table}

\subsection{Entity-Page Lookup for NLR Retrieval}
\label{sparql}
To reduce search cost, the NLR retrieval procedure searches only Wikipedia pages associated with a triple's subject or object entity. This is a retrieval restriction, not a guarantee that all relevant evidence appears on those pages. Entity mappings and the \texttt{dbo:wikiPageID} property are used to obtain page identifiers before searching the Wikipedia dump. The following SPARQL query illustrates this lookup for Freebase entity \texttt{m.03\_r3} (Jamaica), with namespace declarations made explicit. Execution requires a compatible mapping dataset and endpoint.
\lstset{basicstyle=\small\ttfamily,columns=fullflexible,breaklines=true,frame=single,numbers=none,showstringspaces=false}
\begin{lstlisting}
PREFIX owl: <http://www.w3.org/2002/07/owl#>
PREFIX dbo: <http://dbpedia.org/ontology/>
PREFIX fb:  <http://rdf.freebase.com/ns/>
SELECT DISTINCT ?pageID WHERE {
  ?entity owl:sameAs fb:m.03_r3 ;
          dbo:wikiPageID ?pageID .
}
\end{lstlisting}

\subsection{Default Ranking Parameters}
\label{app:default-parameters}
The feature order is $\mathbf{x}=(\mathrm{MS},\mathrm{SL},\mathrm{RR},\mathrm{PT})$. The prior-based ranking rule standardizes each feature using its mean $\mu_i$ and standard deviation $\sigma_i$, then assigns a sign $\tilde w_i\in\{-1,1\}$:
\begin{equation}
\mathrm{QS}(\mathbf{x})=\left[1+\exp\left(-\sum_{i=1}^{4}\tilde w_i\frac{x_i-\mu_i}{\sigma_i}\right)\right]^{-1}
=\left[1+\exp\left(-\sum_{i=1}^{4}\frac{\tilde w_i}{\sigma_i}x_i+\sum_{i=1}^{4}\frac{\tilde w_i\mu_i}{\sigma_i}\right)\right]^{-1}.
\label{qs_}
\end{equation}
Thus the coefficients in Eq.~\eqref{QS} are
\begin{equation}
w_i=\frac{\tilde w_i}{\sigma_i},\label{wi}
\end{equation}
\begin{equation}
b_s=-\sum_{i=1}^{4}\frac{\tilde w_i\mu_i}{\sigma_i}.\label{b_s}
\end{equation}
Setting $\tilde{\mathbf w}=(+1,-1,+1,+1)$ rewards larger $\mathrm{MS}$, $\mathrm{RR}$, and $\mathrm{PT}$, and penalizes greater sentence length. Substituting the statistics in Table~\ref{tab:feature-statistics} gives $w_1=2.224$, $w_2=-0.088$, $w_3=5.392$, $w_4=3.803$, and $b_s=-1.342$ after rounding.
\begin{table}[ht]
\centering
\caption{Feature means and standard deviations.}
\label{tab:feature-statistics}
\begin{tabular}{lrrrr}
\toprule
Statistic & $\mathrm{MS}$ & $\mathrm{SL}$ & $\mathrm{RR}$ & $\mathrm{PT}$ \\
\midrule
$\mu$ & 0.56701 & 18.18729 & 0.08487 & 0.32198 \\
$\sigma$ & 0.44968 & 11.36014 & 0.18546 & 0.26298 \\
\bottomrule
\end{tabular}
\end{table}

\subsection{Experimental Settings}
\label{settings}
Experiments were conducted on a machine with an Intel Xeon Gold 5117 CPU at 2.00 GHz, 256 GB of RAM, and an NVIDIA GeForce RTX 2080 Ti GPU, running Ubuntu 18.04 and CUDA 10.0.

\textbf{GenVer.} The generation model was trained for 300 epochs with learning rate $5\times10^{-6}$, batch size 32, and maximum sequence length 128. The verification model used 100 epochs, learning rate $10^{-5}$, batch size 128, and maximum sequence length 128.

\textbf{KG embedding methods.} TransE, ComplEx, DistMult, TransH, TransR, and TransD were implemented using OpenKE\footnote{\url{https://github.com/thunlp/OpenKE}} \cite{30}. Training used 100 rounds, a learning rate of 0.5, a ranking-loss margin of 5, and an embedding dimension of 100.

\textbf{KGTtm.} The experiments used the released implementation\footnote{\url{https://github.com/TJUNLP/TTMF}} with TransE embeddings on FactFB. The three estimators, TEF, ResourceRank, and RPI, were combined as described by the original method. Training used 200 epochs, an embedding dimension of 100, and dropout of 0.5.

\textbf{\berttriple{}.} This ablation removes the generated NLR and its corresponding feature extractor from GenVer while retaining the other settings.

\textbf{SimKGC.} The experiments used the released implementation\footnote{\url{https://github.com/intfloat/SimKGC}}. The settings are 400 warmup steps, gradient clipping at 10, dropout 0.1 on the final linear layer, and temperature 0.05.

\subsection{Generated NLR Examples}
\label{app:generated-nlrs}
Table~\ref{nlr} shows example triples and generated NLRs for both training-pair configurations. Entity and relation labels are abbreviated for readability; generated text is shown verbatim. The examples show omissions, unsupported details, and changes to relation meaning. They are qualitative examples, not a quantitative evaluation of factual correctness or faithful verbalization.

\small
\setlength{\tabcolsep}{5pt}
\renewcommand{\arraystretch}{1.12}
\begin{longtable}{@{}>{\raggedright\arraybackslash}p{3.4cm}>{\raggedright\arraybackslash}p{5.85cm}>{\raggedright\arraybackslash}p{5.75cm}@{}}
\caption{Generated NLR examples (outputs reproduced verbatim). Score and ML denote training on \scorepairs{} and \mlpairs{}, respectively.}\label{nlr}\\
\toprule
Triple & Generated NLR (Score) & Generated NLR (ML) \\
\midrule
\endfirsthead
\multicolumn{3}{l}{Table~\thetable{} (continued)}\\
\toprule
Triple & Generated NLR (Score) & Generated NLR (ML) \\
\midrule
\endhead
\midrule
\multicolumn{3}{r}{Continued on the next page}\\
\endfoot
\bottomrule
\endlastfoot
S: George Lucas\newline P: Award Nominee\newline O: Djimon Hounsou & Djimon Hounsou was nominated for the George Lucas & Djimon Hounsou was nominated for George Lucas \\
\addlinespace[6pt]
S: Anjelica Huston\newline P: Film Region Date\newline O: Spain & Anjelica Huston said, ``a huge line from Spain to Spain can be found in the us" & Anjelica Huston opened in Spain \\
\addlinespace[6pt]
S: Hugo Award\newline P: Award Category\newline O: Michael Swanwick & Michael Swanwick has received three writers guild of america awards, including nominations for an Hugo Award for Best Novel & Michael Swanwick has the most nominations for one golden globe and Hugo Award for Best Novel \\
\addlinespace[6pt]
S: Scarlett Johansson\newline P: Award Nomination\newline O: Best Female Lead & Scarlett Johansson appeared in the biopic ``pollock" in which she won an Independent Spirit Award for Best Female Lead & Scarlett Johansson won the Independent Spirit Award for Best Female Lead \\
\addlinespace[6pt]
S: Leighton Meester\newline P: Person Profession\newline O: Singer-songwriter & Leighton Meester released the single ``suddenly i feel love" with singer-songwriter & Leighton Meester attended the Professional Children's School and began working as a singer-songwriter \\
\addlinespace[6pt]
S: Actor\newline P: People Profession\newline O: Bono & Bono is the second actor & Bono was such a lousy actor \\
\addlinespace[6pt]
S: conducting\newline P: Performance Actor\newline O: Elmer Bernstein & Elmer Bernstein was surprised by michael aspel at London's royal albert hall after conducting & Elmer Bernstein composed the incidental music score for conducting \\
\addlinespace[6pt]
S: Milla Jovovich\newline P: Film Performance\newline O: Shea Whigham & Shea Whigham met andy robinson, and subsequently started dating until Milla Jovovich & Shea Whigham has appeared in several projects alongside actors Milla Jovovich \\
\addlinespace[6pt]
S: Lincoln\newline P: Nomination Award\newline O: Academy Award & `` Lincoln" was the first film to win both Academy Award for Best Picture and best director & Lincoln won the Academy Award for Best Picture \\
\addlinespace[6pt]
S: Academy Award\newline P: Award Category For\newline O: Guys and Dolls & Guys and Dolls won two golden raspberry awards & Guys and Dolls was nominated for the Academy Award for Best Production Design \\
\addlinespace[6pt]
S: Rob Corddry\newline P: award\_winner\newline O: NBC & Rob Corddry is the only sitcom to have three main episodes in six different networks including the sitcom`` the nutt House" which aired on NBC & Rob Corddry was succeeded by NBC \\
\addlinespace[6pt]
S: comedy-drama\newline P: film\_genre\newline O: B I L B & Bend It Like Beckham is a 1979 american legal comedy-drama & Bend It Like Beckham is an american comedy-drama \\
\addlinespace[6pt]
S: action film\newline P: genre\newline O: The Mummy & The Mummy: Tomb of the Dragon Emperor is a 1978 american-German action film & The Mummy: Tomb of the Dragon Emperor was able to find two universal themes in an action film \\
\addlinespace[6pt]
S: Pursuit of Happyness\newline P: nominated\_work\newline O: Wim Wenders & The Pursuit of Happyness is a 2006 american biographical drama film directed by Wim Wenders & The Pursuit of Happyness is a 2006 american biographical drama film directed by Wim Wenders \\
\addlinespace[6pt]
S: George Clooney\newline P: nominated\_for\newline O: Batman \& Robin & George Clooney was cast because spielberg felt Batman \& Robin & Batman \& Robin is the fourth installment in the``Barack Obama" film series and stars George Clooney \\
\addlinespace[6pt]
S: Yo-Yo Ma\newline P: film\_crew\_gig\newline O: Seeking a Friend & Seeking a Friend for the End of the World is a 2009 experimental documentary film written and directed by Yo-Yo Ma & Seeking a Friend for the End of the World is an american computer-animated musical comedy film directed by Yo-Yo Ma \\
\addlinespace[6pt]
\end{longtable}
\normalsize

\end{document}